\documentclass[letterpaper, 10 pt, journal, twoside]{IEEEtran}

\usepackage{multicol}
\usepackage[bookmarks=true]{hyperref}

\usepackage{float}
\usepackage{amsfonts}
\usepackage{dsfont}
\usepackage{fancyhdr}
\usepackage{comment}
\usepackage{times}
\usepackage{amsmath}
\usepackage{changepage}
\usepackage{amssymb}
\usepackage{graphicx}
\usepackage{adjustbox}
\usepackage{latexsym}
\usepackage{enumerate}
\usepackage[font=small,belowskip=0pt]{caption}
\usepackage{mathtools}
\usepackage{algorithm}
\usepackage{algcompatible}
\usepackage{algpseudocode}
\usepackage{bbm}
\usepackage{xcolor}
\usepackage{bm}
\usepackage{booktabs}

\definecolor{porange}{HTML}{E77500} 
\algrenewcommand{\algorithmiccomment}[1]{\textcolor{porange}{\hfill// #1}}
\algnewcommand{\LineComment}[1]{\Statex \textbf{\textcolor{porange}{// #1}}}

\usepackage{fontawesome5}

\usepackage[%
    style=ieee,
    sorting=none,
    natbib=true,  
    backend=biber,
    sortcites=true,
    doi=false,
    url=false,
    giveninits=true,
    maxcitenames=4, 
    minbibnames=3, 
    maxbibnames=4, 
    hyperref
]{biblatex}

\newbibmacro{string+doiurl}[1]{
    \iffieldundef{doi}{
        \iffieldundef{url}{#1}{\href{\thefield{url}}{#1} }
    }{\href{https://dx.doi.org/\thefield{doi}}{#1}}
}
\DeclareFieldFormat{title}{\usebibmacro{string+doiurl}{\mkbibemph{#1}}}
\DeclareFieldFormat[article,incollection,techreport,inproceedings,book,misc]%
    {title}%
    {\usebibmacro{string+doiurl}{\mkbibquote{#1}}}

\renewbibmacro{in:}{}
\AtEveryBibitem{%
    \clearlist{language}%
    \clearfield{pages}%
    \clearfield{isbn}%
    \clearfield{issn}%
    \clearfield{month}%
    \clearfield{day}
}

\usepackage{balance}

\newtheorem{remark}{Remark}

\newtheorem{proposition}{Proposition}    
    
\newtheorem{definition}{Definition}

\newcommand{\R}{\mathbb{R}}

\newcommand{\dx}{\delta x}

\newcommand{\du}{\delta u}

\DeclareMathOperator*{\argmax}{arg\,max}

\usepackage{ifthen}
\usepackage{xcolor}
\newboolean{include-notes}
\setboolean{include-notes}{false}  
\newcommand{\jaime}[1]{\ifthenelse{\boolean{include-notes}}{\textcolor{orange}{\textbf{Jaime:} #1}}{}}
\newcommand{\haimin}[1]{\ifthenelse{\boolean{include-notes}}{\textcolor{teal}{\textbf{Haimin:} #1}}{}}
\newcommand{\remove}[1]{\ifthenelse{\boolean{include-notes}}{\textcolor{red}{\sout{#1}}}{}}
\newcommand{\new}[1]{\ifthenelse{\boolean{include-notes}}{\textcolor{magenta}{#1}}{#1}}

\newcommand{\shieldset}{\mathcal{S}}

\newcommand{\F}{\mathcal{F}}

\newcommand{\I}{\mathcal{I}}
\newcommand{\bel}{b}
\newcommand{\belspace}{\Delta}
\DeclareMathOperator*{\expectation}{\mathbb{E}}

\newcommand{\bX}{\mathbb{X}}

\newcommand{\ctrlset}{\mathcal{U}}

\newcommand{\cN}{\mathcal{N}}
\newcommand{\cL}{\mathcal{L}}

\newcommand{\br}{\text{br}}
\newcommand{\tn}{{\tilde{n}}}
\newcommand{\bn}{\mathbf{n}}

\newcommand{\lat}{\text{lat}}

\renewcommand{\int}{\operatorname{int}}

\newcommand{\pre}{\operatorname{pre}}

\newcommand{\QMDP}{\text{QMDP}}

\newcommand{\cl}{\text{cl}}
\newcommand{\Sim}{\text{sim}}
\newcommand{\shield}{\text{\tiny{\faShield*}}}

\begin{document}
%
\title{SHARP: Shielding-Aware Robust Planning for Safe and Efficient Human-Robot Interaction}
%
%
%

\author{Haimin Hu$^{1}$, Kensuke Nakamura$^{2}$, and Jaime F. Fisac$^{1}$%
\thanks{Manuscript received: September 9, 2021; accepted February 6, 2022. This paper was recommended for publication by Associate Editor Gentiane Venture and Editor Tamim Asfour upon evaluation of the reviewers' comments.
This work was supported by Princeton University's Project X Program.} 
\thanks{$^{1}$Department of Electrical and Computer Engineering, Princeton University, {\tt\footnotesize \{haiminh,jfisac\}@princeton.edu}}%
\thanks{$^{2} $Department of Mechanical and Aerospace Engineering, Princeton University, {\tt\footnotesize k.nakamura@princeton.edu}}%
\thanks{Digital Object Identifier (DOI): see top of this page.}
}
%
%

\markboth{IEEE Robotics and Automation Letters. Preprint Version. February, 2022}
{Hu \MakeLowercase{\textit{et al.}}: Shielding-Aware Robust Planning for Safe and Efficient Human-Robot Interaction} 

%


\maketitle

\begin{abstract}
Jointly achieving safety and efficiency in \new{human-robot interaction} settings is a challenging problem, as the robot's planning objectives may be at odds with the human's own intent and expectations.
Recent approaches ensure safe robot operation in uncertain environments through a supervisory control scheme, sometimes called ``shielding'', which overrides the robot's nominal plan with a safety fallback strategy when a safety-critical event is imminent.
These reactive ``last-resort'' strategies (typically in the form of aggressive emergency maneuvers) focus on preserving safety without efficiency considerations;
when the nominal planner is unaware of possible safety overrides, shielding can be activated more frequently than necessary, leading to degraded performance.
In this work, we propose a new shielding-based planning approach that allows the robot to plan efficiently by explicitly accounting for possible future shielding events.
Leveraging recent work on Bayesian human motion prediction, the resulting robot policy proactively balances nominal performance with the risk of high-cost emergency maneuvers triggered by low-probability human behaviors.
We formalize Shielding-Aware Robust Planning (SHARP) as a stochastic optimal control problem and propose a computationally efficient framework for finding tractable approximate solutions at runtime.
Our method outperforms the shielding-agnostic motion planning baseline (equipped with the same human intent inference scheme) on simulated driving examples with human trajectories taken from the recently released Waymo Open Motion Dataset.
\end{abstract}


%
\IEEEpeerreviewmaketitle

\section{Introduction}
%
%
%
%
\IEEEPARstart{I}{n} recent years, much effort has been devoted to developing robotic systems that can coexist and interact with humans.
Indeed, in order to serve people in daily life, autonomous systems must competently predict and seamlessly adapt to human behaviour.
Examples include autonomous driving~\cite{sadigh2018planning,fisac2019hierarchical}, indoor aerial robots~\cite{fisac2018probabilistically} and robotic arms~\cite{amor2014interaction}.
These applications are safety-critical, since inappropriate robot behaviours can pose significant danger to humans.
Therefore, it is crucial to develop motion planning algorithms for \new{human-robot interaction} that not only yield high performance but also guarantee safety at all times.

\begin{figure}[!hbtp]
  \centering
  \includegraphics[width=1.0\columnwidth]{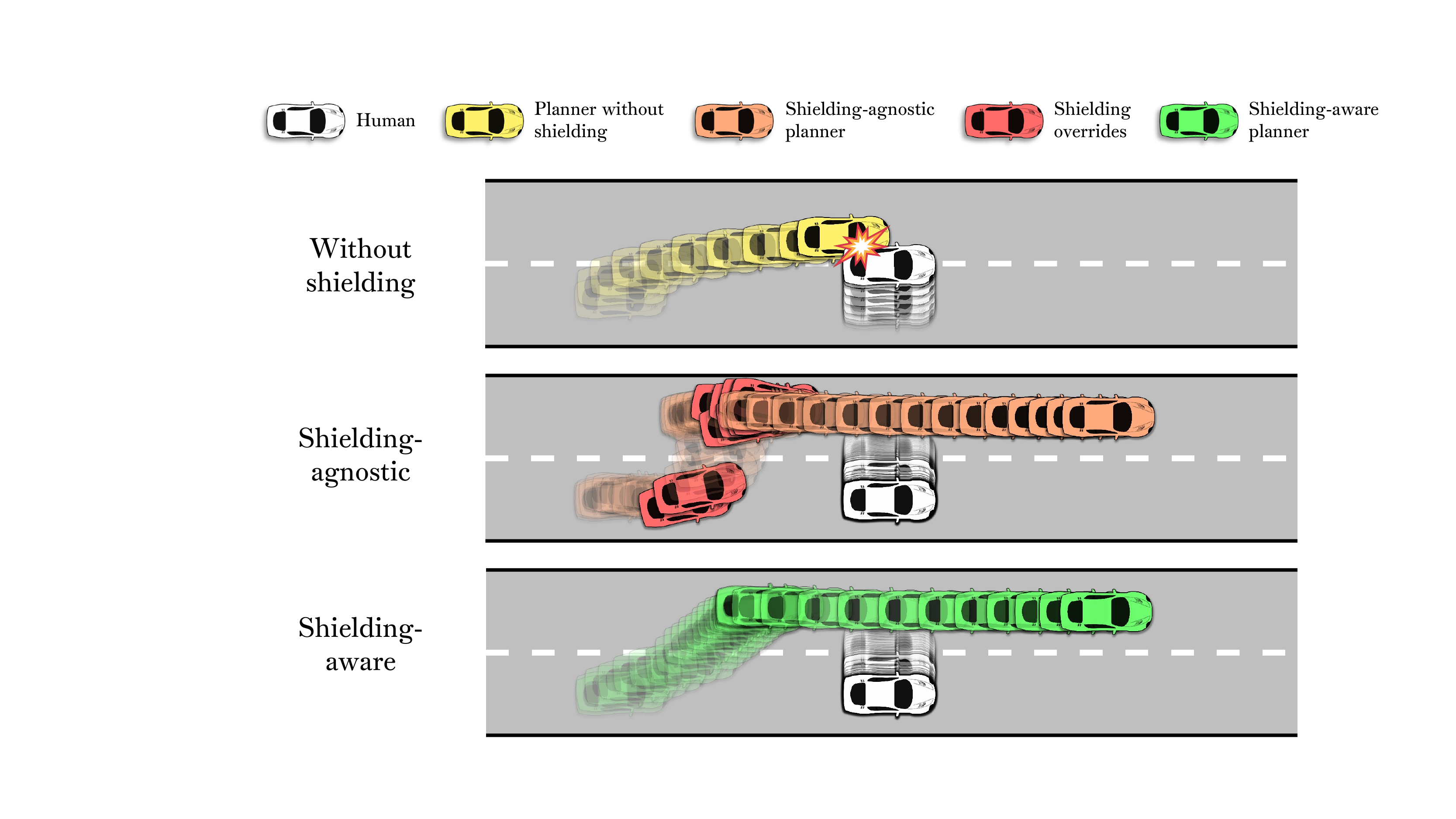}
  \caption{\label{fig:cover} \small 
  An autonomous car seeks to overtake a vehicle driven by a distracted human (longitudinal positions are shown in relative coordinates).
  \textit{Top:} A planner without formal safety guarantees incurs in a collision.
  \textit{Middle:} A shielding-agnostic planner triggers safety overrides unnecessarily often, maintaining safety at the cost of performance and comfort.
  \textit{Bottom:} Our proposed planner reasons about future shielding events and avoids relying on safety overrides if possible, which significantly improves the resulting performance.
  }
\end{figure}

In typical \new{human-robot interaction} scenarios, since the robot's safety and performance are naturally coupled with the human's movements, the robot must be able to make real-time inferences about the human's future motion during planning. 
Predicting human motion while planning the robot's trajectory can be generally cast as a partially-observable stochastic game~\cite{hansen2004dynamic}.
In~\cite{fisac2019hierarchical}, the authors modeled the interaction between the human and the robot as a dynamic game that allows for real-time trajectory planning.
In~\cite{sadigh2018planning}, the authors simplified the problem to an open-loop Stackelberg game and showed that the human's objective function can be 
learned using inverse reinforcement learning methods~\cite{ziebart2008maximum}.

Comparing to the large body of work on performance-oriented planning for \new{human-robot interaction}, ensuring safe interactions subject to uncertain human motion is a relatively less explored topic.
One popular way of achieving safety for \new{human-robot interaction} tasks is by adding to the planning problem a chance constraint or cost penalty for collision avoidance, which is then accounted for via probabilistic predictions of the human's future motion
(see for example~\cite{sadigh2018planning}).
In~\cite{fisac2018probabilistically}, the authors proposed to have the robot maintain a runtime measure of its degree of confidence in a learned human model.
This allows the robot to plan probabilistically safe trajectories accounting for the observed accuracy of its own human motion predictions.
Ultimately, however, under any such probabilistic approaches, safety can be compromised when the human takes low-probability actions.
This is also known as the issue of the ``long tail'' of unlikely events~\cite{koopman2018heavy}.

In general, all-time safety in \new{human-robot interaction} can be ensured by a least-restrictive supervisory control scheme, often referred to as \textit{shielding}.
This approach involves synthesizing and implementing a reactive safety fallback policy as the ``last-resort'', which overrides a nominal policy \textit{only when} a safety-critical event, e.g. a collision, is imminent.
Such shielding mechanisms include, for example, reachability analysis~\cite{bansal2017hamilton,zhang2021safe,hsu2021safety}, control barrier functions~\cite{ames2016control,robey2020learning}, Lyapunov methods~\cite{chow2018lyapunov}, and model predictive {control}~\cite{li2020safe,wabersich2021predictive}.
Despite being effective at guaranteeing safety, applying shielding too frequently can greatly degrade the planning performance of the robot, since the safety controllers are typically designed without performance consideration such as task completion time, passenger comfort or energy consumption.

Simultaneously ensuring safety and optimizing performance for human-robot interaction tasks can be formulated as a stochastic optimal control problem (OCP), which combines propagating uncertainty (i.e. human motion), guaranteeing safety and optimizing the robot's objectives altogether in a single optimization problem.
In principle, a stochastic OCP can be solved using stochastic dynamic programming~\cite{bertsekas1995dynamic}, which is, however, only tractable for toy examples~\cite{klenske2016dual}.
Recent work~\cite{arcari2020approximate} proposes to approximately solve the OCP using stochastic model predictive control (SMPC) methods~\cite{bernardini2011stabilizing}.

\textbf{Statement of contributions:} 
In this paper, we propose a novel shielding-aware planning framework that jointly achieves safety and performance for human-robot interaction.
The key element of our approach is the formulation of a stochastic OCP that reasons about future shielding events via human motion prediction, while optimizing the robot's trajectory.
The resulting policy improves the planning performance by preventing the robot from having to apply a costly shielding maneuver \emph{in the future}.
We reformulate the OCP by exploiting the structure in the human uncertainty model and solve it using efficient approximate dynamic programming methods.
We evaluated our approach on simulated driving scenarios, with the human driver's trajectories taken from the Waymo Open Motion Dataset~\cite{sun2020scalability}.
On average, our proposed planner improved the planning performance by at least $16\%$ comparing to the state-of-the-art SMPC baseline across all testing scenarios.


\section{Preliminaries}

\subsection{Dynamical Systems}
We consider a broad class of discrete-time dynamical systems for the robot and human, respectively,
\begin{equation}
\label{eq:subsys_R_H}
x^R_{t+1} = f^R (x_t^R, u_t^R), \quad x^H_{t+1} = f^H (x_t^H, u_t^H),
\end{equation}
where the input constraints are $u_t^R \in \ctrlset^R \subseteq \R^{m_R}$ and $u_t^H \in \ctrlset^H \subseteq \R^{m_H}$.
We now define a joint system that captures the interactions between the human and robot subsystems,
\begin{equation}
\label{eq:joint_sys}
x_{t+1} = f (x_t, u^R_t, u^H_t),
\end{equation}
where $f: \R^{n_x} \times \ctrlset^R \times \ctrlset^H \rightarrow \R^{n_x}$ are the joint human-robot dynamics, whose state vector is given by $x_t = \Phi \left[x_t^R, x_t^H \right]$ and $\Phi: \R^{n_x} \times \R^{n_{R}+n_{H}}$ is a change-of-coordinates matrix.

\begin{remark}
{The theoretical analysis in this paper extends to multi-human interaction by letting $x_t^H, u_t^H$ in \eqref{eq:joint_sys} represent the \emph{joint} state and actions of multiple humans.
Computational scalability is limited in practice by the
exponential complexity common to combinatorial problems of this kind.
}
\end{remark}
\textbf{Running example:} 
We consider a highway driving scenario, as depicted in Fig.~\ref{fig:cover}, involving an autonomous vehicle~($R$) and a human-driven vehicle ($H$), each modeled by {simplified dynamics} taken from~\cite{fisac2019hierarchical}.
The states are the relative longitudinal position $p_x^r$, relative velocity $v_r$ and lateral positions $p_y^i, \ i \in \{R,H\}$.
The controls are the desired lateral velocity $v_{\lat}^i$ and acceleration $a^i$.
The robot's task is to safely overtake the human.

\subsection{Safe Human-Robot Interactive Planning via Shielding}
In this paper, we focus on safety-critical human-robot interaction applications in which the state trajectory of the human-robot joint system must not enter a failure set $\F \subseteq \R^{n_x}$.
This includes, for example, the robot colliding with the human.
To ensure that $x_t \notin \F$ for all $t \geq 0$ despite the \textit{worst-case} human actions, we make use of a supervisory safe control strategy, often referred to as ``shielding'', which is defined as a tuple $(\Omega, \pi^s)$.
Here, set $\Omega \subseteq \R^{n_x}$ is a \textit{safe set} that satisfies $\Omega \cap \F = \emptyset$, and $\pi^s: \R^{n_x} \rightarrow \ctrlset^R$ is a safe control {policy} that keeps the state inside $\Omega$ even under the worst-case human action.
This is formalized in the following definition.
\begin{definition}[Robust controlled-invariant set]
\label{def:RCI_set}
{Given dynamics~\eqref{eq:joint_sys} with a bounded uncertain input $u_t^H\in\ctrlset^H$,}
a set ${\Omega \subseteq \R^{n_x}}$ is a robust controlled-invariant set {if there exists a control policy ${\pi^s: \R^{n_x}\! \rightarrow \ctrlset^R}$ that keeps $x_t$ from leaving~$\Omega$}:
\end{definition}
\begin{equation}
    \vphantom{\bigg(}
    x_0 \in \Omega \Rightarrow x_t \in \Omega, \ \forall t>0, \ \forall u^H_t \in \ctrlset^H, \ {u^R_t=\pi^s(x_t)}.
\end{equation}
\looseness=-1
The definition {suggests} that the safe control {$\pi^s(x_t)$} is needed only when the state is \emph{about to leave} the safe set.
{
Let the \emph{shielding set} $\shieldset^R\subset\Omega\times\ctrlset^R$ contain all state-action pairs that \emph{might} result in the next state being outside of the safe set:
\begin{equation}
\label{eq:shield_set}
    \shieldset^R = \{ (x,u^R) \in \Omega\times\ctrlset^R \mid \exists \tilde{u}^{H} \!\! \in \ctrlset^{H}\!\!: f\left(x, u^R, \tilde{u}^{H}\right) \notin \Omega \}.
\end{equation}
}
{We can then define a ``least-restrictive'' supervisory safety filter 
for arbitrary
candidate control actions $\tilde{u}^R_t$:
\begin{equation}
\label{eq:shield_control}
u^R_{t}=
\pi^\shield(x_t; \tilde{u}^R_t)\!:=
\begin{cases}
\tilde{u}^R_t,  & \text{if } (x_t, \tilde{u}^R_t) \not\in \shieldset^R \\
\pi^s(x_t),  & \text{if } (x_t, \tilde{u}^R_t) \in \shieldset^R
\end{cases}
\end{equation}
}
{The \emph{shielding mechanism}} \eqref{eq:shield_control} allows the robot to apply \textit{any} nominal controller $\pi_t: \R^{n_x}\! \rightarrow \ctrlset^R$ as long as
{$\big(x_t,\pi_t(x_t)\big)$ is not in the shielding set $\shieldset^R$;
otherwise, it overrides $\pi_t(x_t)$ with the safety policy $\pi^s(x_t)$}.
The result below follows.
\begin{proposition}[Shielding]
\label{prop:shield}
If a set $\Omega$ is robust controlled-invariant under $\pi^s(\cdot)$, then it is robust controlled-invariant under
{$\pi^\shield\big(\,\cdot\,;\pi_t(\cdot)\big)$, for}
any nominal {control policy} $\pi_t(\cdot)$.
\end{proposition}

{Equation~\eqref{eq:shield_control} and Proposition~\ref{prop:shield}} describe a variety of shielding mechanisms, {from Hamilton-Jacobi and Lyapunov analysis~\cite{bansal2017hamilton,chow2018lyapunov} to predictive policy \mbox{rollouts~\cite{zhang2021safe,li2020safe,wabersich2021predictive}}}.
In this paper, we focus on efficient {shielding-aware} planning,
only assuming that we have access to \emph{some} shielding mechanism 
{$\pi^\shield$}.
Our framework is therefore quite general and can work in conjunction with many existing shielding methods.


\textbf{Running example:}
A typical failure set for system~\eqref{eq:joint_sys}~is
{${\F := \{ x \in \R^{4} \mid \big(| p_x^r | < 5.5 \text{ m} \land |p_y^R - p_y^H| < 2.0 \text{ m}\big)} \lor |p_y^R| > 3.7 \text{ m} \}$, including any loss of separation between the two vehicles as well as $R$ exceeding the road edges. Note that $\F$ is a static set in the joint state space, even as $H$ and $R$ move.}
We use Hamilton-Jacobi (HJ) reachability~\cite{bansal2017hamilton} to compute the safe set $\Omega$ and control {policy} $\pi^s$ for shielding.

\subsection{Predicting Human Motion}
\label{subsec:predict_H}
The robot's main task is to achieve desirable performance through minimizing a cost function $\ell^R(x_t, u_t^R)$ over time.
Note that both the cost function and the safe controller $\pi^s(x_t)$ depend on human's state $x^H_t$.
Therefore, in order to plan efficiently, the robot must be able to predict the human's actions, since they can not only affect the robot's cost directly, but also indirectly by triggering (costly) shielding events.
Here, we use the ``noisily-rational'' Boltzmann model originated from cognitive science~\cite{luce1959individual} to predict human's future motion.
Concretely, the probability of $H$ taking a specific action $u^H_t \in \ctrlset^H$ is given by,
\begin{equation}
\label{eq:Boltzmann}
    P\left(u_{t}^{H} \mid x_{t}, \beta_{t}, \theta_{t}\right)=\frac{e^{-\beta_{t} Q_{\theta_{t}}^{H}\left(x_{t}, u_{t}^{H}\right)}}{\sum_{\tilde{u}_{t}^{H} \in \tilde{\ctrlset}^{H}} e^{-\beta_{t} Q_{\theta_{t}}^{H}\left(x_{t}, \tilde{u}_{t}^{H}\right)} },
\end{equation}
where $Q_{\theta_t}^H(x_t, u^H_t)$ is the human's state-action value function, characterized by a set of time-varying parameters $\theta_t \in \R^{n_{\theta}}$ indicating human's possible intents.
The inverse temperature ${\beta_t > 0}$, sometimes called ``rationality coefficient'' or ``model confidence'', quantifies the tendency of the human's actions to concentrate around the modeled optimum.
This model assumes that the human is exponentially likelier to pick actions with better state-action values.

\begin{remark}
Our framework is agnostic to the concrete methods for determining the human's possible intents $\theta$, which is usually specified by the system designer based on domain knowledge or learned from prior data.
Goal-driven models of human motion are well-established in the literature. See for example~\cite{sadigh2018planning,ziebart2008maximum}.
\end{remark}

\textbf{Running example:} 
The human's state-action value function is expressed as the convex combination of two basis functions, $Q_{\theta_t}^H(\cdot) = \theta_t Q^H_1(\cdot) + (1-\theta_t) Q^H_2(\cdot), \theta_t \in [0,1]$,
where $Q^H_1(\cdot)$ and $Q^H_2(\cdot)$ are quadratic functions capturing $H$ tracking two possible intents: driving in the left and right lane, respectively, at the cruising speed $30$ m/s.

\subsection{Inferring Human Model Parameters}
In general, parameters $(\beta_t,\theta_t) \in \Xi \subseteq \R_{\geq 0} \times \R^{n_{\theta}}$ at each time instance $t$ are unknown to the robot and therefore can only be estimated from past observations.
To address this, we define the information vector $\I_t := \left[x_{t}, u_{t-1}^{H}, \mathcal{I}_{t-1}\right]$ as the collection of all \textit{causally observable} information at time $t \geq 0$, with $\mathcal{I}_{0}=\left[x_{0}\right]$.
We then define the \textit{belief state} ${\bel_t := P\left(\beta_{t}, \theta_{t} \mid \mathcal{I}_{t}\right)}\in\belspace$ as the probability distribution of parameters $(\beta_t,\theta_t)$ conditioned on $\I_t$, and $\bel_0:=P\left(\beta_{0}, \theta_{0} \right)$ is a given prior distribution.
When the robot receives a new observation $u_{t}^{H} \in \I_{t+1}$, the current belief state $\bel_t\in\belspace$ is updated using the recursive Bayesian estimation,
\begin{align}
\label{eq:Bayesian_est}
\bel^-_{t+1} &:= P(\beta_{t}, \theta_{t} \mid \mathcal{I}_{t+1}) \notag \\
&= \frac{P(u^H_t \mid x_t, \beta_t, \theta_t) \bel_{t}(\beta_t, \theta_t) }{\sum_{(\tilde{\beta}, \tilde{\theta}) \in \tilde{\Xi}} P(u^H_t \mid x_t, \tilde{\beta}, \tilde{\theta}) \bel_{t}(\tilde{\beta}, \tilde{\theta}) }
\end{align}
\begin{figure}[!hbtp]
  \centering
  \includegraphics[width=1.0\columnwidth]{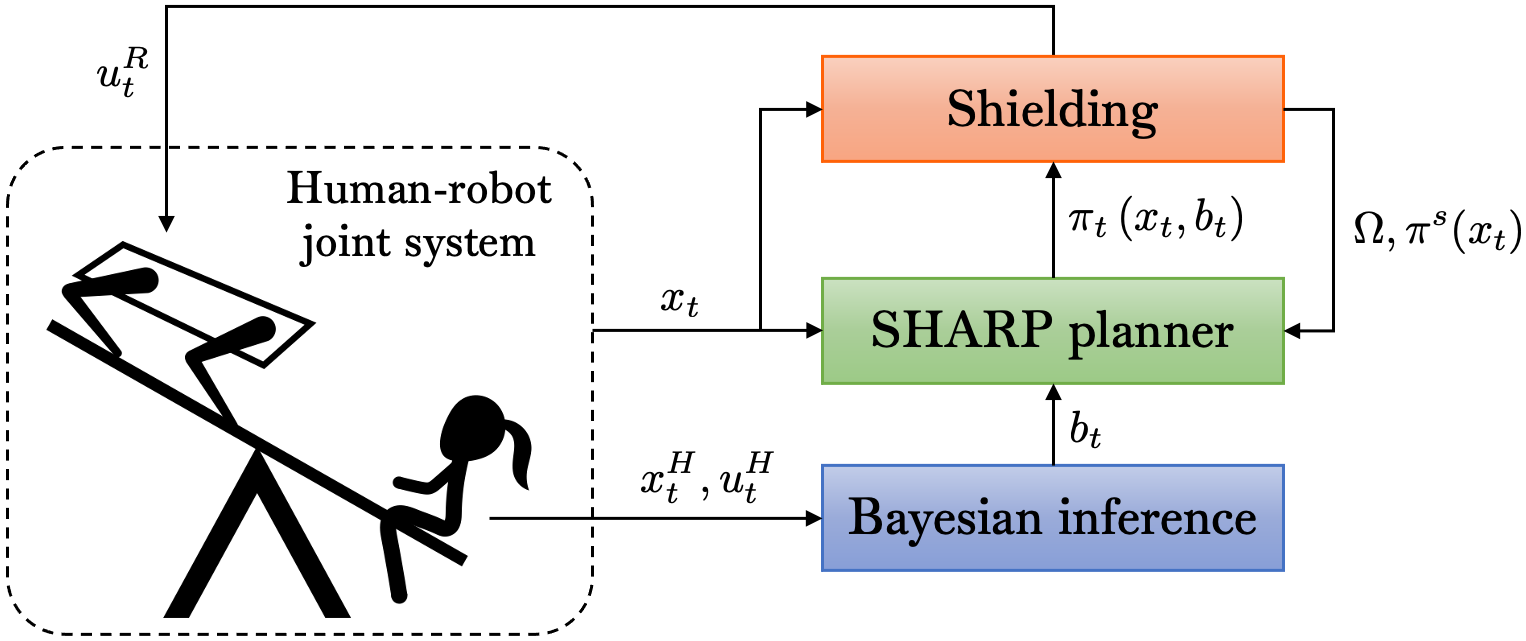}
  \caption{\label{fig:framework} Overview of the proposed SHARP framework.
  }
\end{figure}
\begin{align}
\label{eq:Bayesian_est_trans}
\bel_{t+1}
&= P(\beta_{t+1}, \theta_{t+1} \mid \mathcal{I}_{t+1}) \notag \\ &= \textstyle \sum_{(\tilde{\beta}, \tilde{\theta}) \in \tilde{\Xi}}  P(\beta_{t+1}, \theta_{t+1} \mid \tilde{\beta}, \tilde{\theta}) P(\tilde{\beta}, \tilde{\theta} \mid \mathcal{I}_{t+1})
\end{align}
where $P(\beta^\prime, \theta^\prime \mid \beta, \theta)$ is a transition model and a set $\tilde{\Xi}$ discretized from $\Xi$ is used as the support.
We can then rewrite~\eqref{eq:Bayesian_est} and~\eqref{eq:Bayesian_est_trans} compactly as a dynamical system,
\begin{equation}
\label{eq:belief_state_dyn}
{\bel}_{t+1}=g({\bel}_{t}, {x}_{t}, {u}^H_{t} ).
\end{equation}


\section{SHARP: Shielding-Aware Robust Planning}
In this paper, our goal is to plan an efficient trajectory for the robot while ensuring safety at all times.
A na\"ive approach would be using a shielding-agnostic nominal planner in \eqref{eq:shield_control}, whose main focus is on performance but is unaware of the possibility of being overridden by the shielding mechanism.
This approach can, however, yield a trajectory far from optimal in the presence of noisily rational human agents.
The main reason is that shielding-agnostic planners tend to \emph{unwittingly} activate shielding, resulting in frequent discrepancies between the efficiently planned trajectory, which will not be allowed to take place, and the costly executed trajectory, which was unaccounted for in planning. 
Conversely, a planner with shielding awareness reasons about potential future shielding events based on human motion predictions and preempts unnecessary overrides, thereby improving closed-loop performance.

Based on this central insight, we propose a new planning formulation that accounts for possible future shielding events, which we call Shielding-Aware Robust Planning (SHARP).
The core of SHARP is a stochastic optimal control problem formulated as follows:
\begin{subequations}
\label{eq:SHARP}
\begin{align}
\label{eq:SHARP:obj} \min _{\substack{\pi_{[0:N-1]}}} \
&\operatorname*{\mathbb{E}}\limits_{\substack{ \beta_{[0:N-1]}, \theta_{[0:N-1]}, \\ {u}_{[0: N-1]}^{H} } } \sum_{k=0}^{N-1} \ell^R (x_k, u^R_k) + \ell_F^R (x_N) \\
\text{s.t.} \qquad &x_{0}=x_{t}, \ \bel_{0}=\bel_{t}, \label{eq:SHARP:sys_init}\\
& \forall k = 0,\ldots,N-1: \notag \\
\label{eq:SHARP:sys_dyn} &{x}_{k+1}={f}\left({x}_{k}, {u}^R_{k}, {u}^H_{k}\right) \\
\label{eq:SHARP:belief_dyn} &{\bel}_{k+1}=g\left({\bel}_{k}, {x}_{k}, {u}^H_{k}\right) \\
\label{eq:SHARP:control}
&u^R_{k}=
    {\pi^\shield\big(x_t; \pi_k(x_k, \bel_k)\big)}
\end{align}
\end{subequations}
where $\ell^R: \R^{n_x} \times \ctrlset^R \rightarrow \R_{\geq 0}$ and  $\ell^R_F: \R^{n_x} \rightarrow \R_{\geq 0}$ are designer-specified stage and terminal cost function, and ${\pi_k : \R^{n_x}\times\belspace \to \ctrlset^R}$ is a \emph{causal} feedback policy that leverages the (yet-to-be-acquired) knowledge of $x_k$ and $\bel_k$.

In theory, problem \eqref{eq:SHARP} can be solved using stochastic dynamic programming~\cite{bertsekas1995dynamic}.
An optimal value function $V_k(x_k, \bel_k)$ and control policy $\pi_k^*(x_k, \bel_k)$ can be obtained backwards in time using the Bellman recursion,
\begin{equation}
\label{eq:sto_DP}
\begin{aligned}
V_k(x_k, \bel_k) = &\min_{\substack{
\pi_k(x_k,\bel_k)}} \ell^R(x_k, u^R_k) \\
+ & \operatorname*{\mathbb{E}}\limits_{\substack{ (\beta_{k}, \theta_{k}) \sim \bel_k, {u}_{k}^{H} } } \left[V_{k+1}(x_{k+1}, \bel_{k+1}) \mid \mathcal{I}_k \right] \\
& \quad \text{s.t.} \ \ \eqref{eq:SHARP:sys_dyn}-\eqref{eq:SHARP:control}
\end{aligned}
\end{equation}
with terminal condition $V_N(x_N,\bel_N) = \ell^R_F(x_N)$.
Due to causal feedback, the controller obtained by solving \eqref{eq:sto_DP} takes into account information that will become available in the future.
As a result, the robot is able to predict upcoming shielding events using not only the \emph{current} belief state $\bel_t$, but also a series of potential \emph{future} belief states propagated via \eqref{eq:SHARP:belief_dyn}, thus gaining an opportunity to plan a more efficient trajectory while staying safe \emph{without} relying on the (usually) costly shielding maneuvers.
Unfortunately, \eqref{eq:sto_DP} is computationally intractable in all but the simplest cases.
Even with spatial discretization, the belief states $\bel_k$ generally live in a high dimensional space, which makes solving \eqref{eq:sto_DP} infeasible in practice due to the ``curse of dimensionality''~\cite{bertsekas1995dynamic}.

Next, we focus on developing a tractable and efficient computation framework for solving OCP \eqref{eq:SHARP} approximately.
Our road map is to reformulate \eqref{eq:SHARP} in two ways, each tackled with a different approximate dynamic programming method.
Our main focus is on reformulating \eqref{eq:SHARP} as a scenario-tree-based stochastic model predictive control (ST-SMPC) problem, which is a real-time trajectory optimization method originally developed in~\cite{bernardini2011stabilizing}.
This approach estimates the expectation in \eqref{eq:SHARP:obj} and propagates the belief states in \eqref{eq:SHARP:belief_dyn} based on a small number of likely uncertainty realizations, thereby preserving a simplified but representative truncation of the original problem's structure.
However, we first present a simpler relaxation of~\eqref{eq:sto_DP} based on the QMDP assumption~\cite{littman1995learning}, which allows computing a tabular solution offline.
The solution is a value function that approximately captures the cost-to-go over the full horizon $N$, and can be used as a guiding terminal cost function in ST-SMPC to implicitly extend the planning horizon.
The overall SHARP framework is illustrated in Fig.~\ref{fig:framework}.

\subsection{Problem Simplification with the QMDP Assumption}
\label{sec:SHARP:ADP}

In this section, we discuss how to solve a relaxation of~\eqref{eq:sto_DP} with an offline tabular dynamic programming scheme.
We start by discretizing the joint state space and robot's action space into $\tilde{\bX}$, $\tilde{\ctrlset}^R$, and letting $z_t := [x_t, \beta_t, \theta_t]$.
Now, under perfect observability of $z_t$,
we would have a fully certain belief ${b_t\equiv\mathds{1}_{(\beta_t,\theta_t)}}$
and \eqref{eq:sto_DP} would reduce to a full-information problem that can be numerically solved with the Bellman recursion:
\begin{equation}
\label{eq:tab_DP}
\begin{aligned}
\tilde{V}_k(z_k) = &\min_{\substack{
\pi_k(z_k)}} \ell^R(x_k, u^R_k) \\
+ & \sum_{(\tilde{\beta}, \tilde{\theta}) \in \tilde{\Xi}} P( \tilde{\beta}, \tilde{\theta} \mid \beta_{k}, \theta_{k}) \expectation_{\substack{{u}_{k}^{H} } } \left[\tilde{V}_{k+1}(\tilde{z}_{k+1}) \right] \\
&\quad \text{s.t.} \ \ \eqref{eq:SHARP:sys_dyn}, \  \eqref{eq:SHARP:control},
\end{aligned}
\end{equation}
where $\tilde{z}_{k+1} := [x_{k+1}, \tilde{\beta}, \tilde{\theta}]$.
This simplified Bellman recursion follows the QMDP assumption~\cite{littman1995learning}, which optimistically assumes that the uncertainties in the current belief states $(\beta,\theta)$ disappear in one time-step.
Here, in lieu of evolving the belief states with the measurement update~\eqref{eq:Bayesian_est}, uncertainties in $(\beta,\theta)$ are now propagated only by the transition model $P(\beta^\prime, \theta^\prime \mid \beta, \theta)$ in \eqref{eq:Bayesian_est_trans}.
As a result, the Bellman recursion \eqref{eq:tab_DP} can be computed efficiently, at the cost of losing the ability to account for future uncertainties.
Given a state $x_t$, a belief state $b_t$ and a lookup table of $\tilde{V}_0(\cdot)$ obtained by \eqref{eq:tab_DP}, we can obtain a value function,
\begin{equation}
\label{eq:QMDP_ter_cost}
\begin{aligned}
V_F(x_t,b_t) :=&
\min_{\substack{\pi_k(x_t,b_t)}} \ell^R(x_t, u^R_t)\\ &+\expectation_{\substack{ (\beta_t,\theta_t) \sim b_t }}
\expectation_{\substack{{u}_{k}^{H} } }
\expectation_{(\tilde{\beta}, \tilde{\theta})} 
\left[\tilde{V}_{0}(\tilde{z}_{t+1}) \right],
\end{aligned}
\end{equation}
which is an optimistic estimate of the true cost-to-go $V_t(x_t,b_t)$ of \eqref{eq:sto_DP}.
In Section \ref{sec:SHARP:ST-SMPC}, we will use this approximate value function as a guiding terminal cost in ST-SMPC.
As a byproduct of \eqref{eq:QMDP_ter_cost}, we can obtain a causal feedback control policy, which we refer to as SHARP-QMDP.
In the next section, we will use this policy to construct a scenario tree for ST-SMPC.
Nonetheless, it can also be used directly as the nominal planner in \eqref{eq:shield_control} for online planning.
Although this policy no longer propagates belief states, it is still effective at predicting shielding events and gains an information advantage over a shielding-unaware policy due to causal feedback and the shielding constraint \eqref{eq:SHARP:control}.

\subsection{ST-SMPC with the Sparse LQG Tree}
\label{sec:SHARP:tree}

The performance of SHARP-QMDP can be limited by its inability to propagate the belief states with measurements on human uncertainties.
In this section, we focus on developing a shielding-aware planner that propagates the belief states and leverages them to better predict future shielding events.
Motivated by recent advances in approximate dynamic programming for uncertain systems~\cite{arcari2020approximate,bonzanini2020safe}, we propose to propagate the belief states in \eqref{eq:SHARP:belief_dyn} using samples of $u^H_k$.
This leads to a scenario tree that allows us to reformulate \eqref{eq:SHARP} as a computationally tractable ST-SMPC problem.
With discretized human action and parameter spaces $\tilde\ctrlset^H,\tilde\Xi$,
the (intractable) Bellman recursion \eqref{eq:sto_DP} can be evaluated for any given state $x_0$ and belief state $b_0$:
\begin{equation}
\label{eq:exp_expand}
\begin{aligned}
V_0(x_0, \bel_0) =&\min_{\substack{
u_0 \in \ctrlset^{R}}} \ell^R(x_0, u^R_0) + \textstyle\sum_{\beta, \theta} b_0(\beta, \theta) \cdot \\
&\textstyle\sum_{\tilde{u}^H_0 \in \tilde{\ctrlset}^H}  P(\tilde{u}^{H} \mid x^H_{0}, \beta, \theta ) V_{1}(\tilde{x}_{1}, \tilde{\bel}_{1}),
\end{aligned}
\end{equation}
with value functions at subsequent times obtained recursively in an analogous manner.
The next state $\tilde{x}_{1}$ and belief state ${\tilde{\bel}_{1}}$ are obtained by computing ${\tilde{x}_{1}={f}\left(\pre(\tilde{x}_{1}), \tilde{u}^{R}_{0}, \tilde{u}^{H}_{0}\right)}$ and $\tilde{\bel}_{1}=g(\pre(\tilde{\bel}_{1}), \pre(\tilde{x}^H_{1}), \tilde{u}^{H}_{0})$.
\begin{figure}[!hbtp]
  \centering
  \includegraphics[width=1.0\columnwidth]{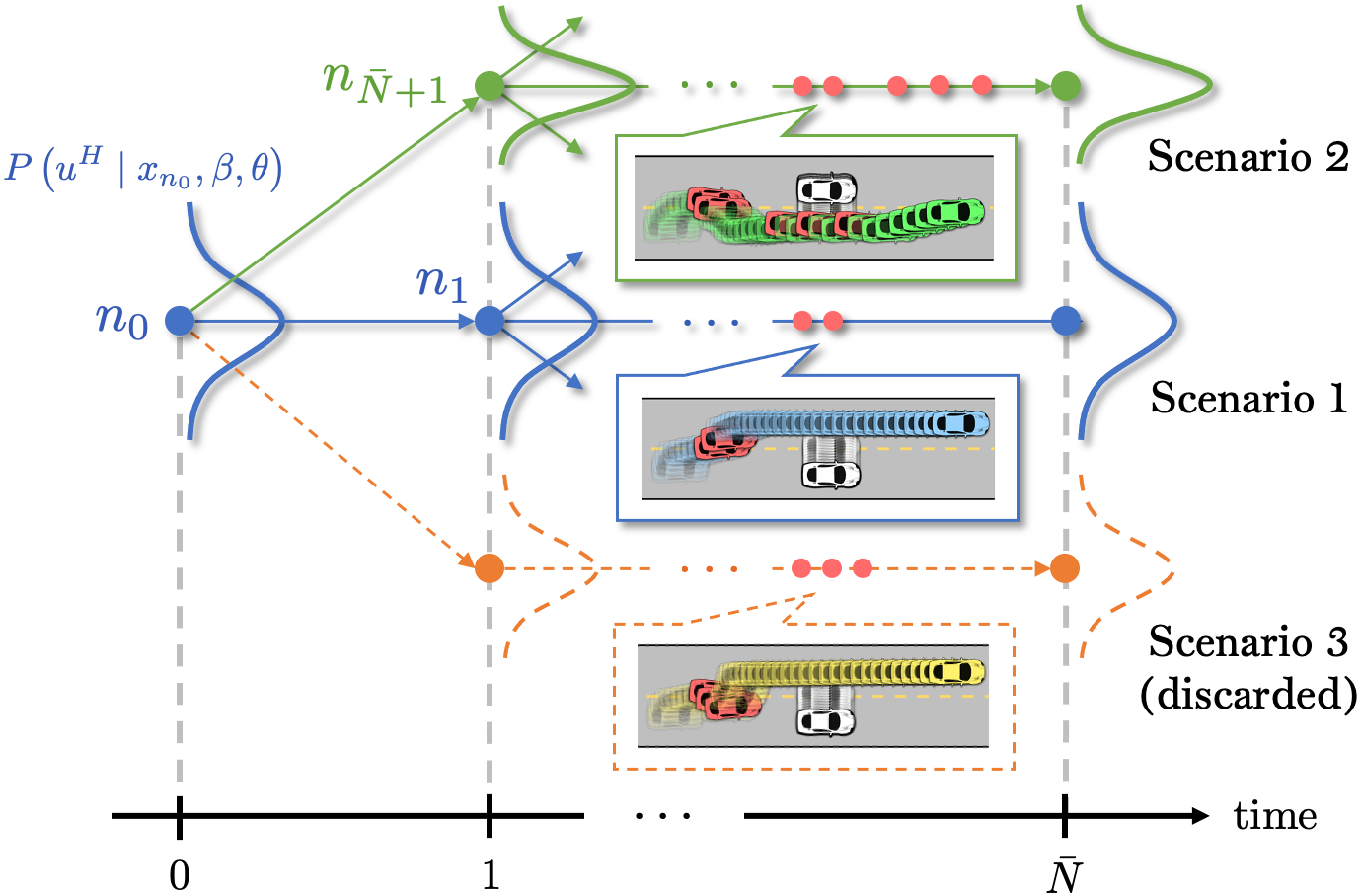}
  \caption{\label{fig:sparse_tree} Illustration of a sparse LQG scenario tree. Red (colored) dots denote (non-)shielding nodes. The bell curve at node $\tn$ represents the Gaussian distribution $P(u^H \mid x_{\tn}, \beta, \theta)$. Two scenarios (1 and 2) are branched out from the root node $n_0$, while Scenario 3 is discarded due to its similarity with Scenario 1.
  }
\end{figure}
\noindent Here, ${\pre(\tilde{x}_{1}) := x_0}$ is the predecessor state of $\tilde{x}_{1}$, similarly for beliefs.
Given a sequence of human uncertainty realizations $(\tilde{\beta}_{[0:N-1]}, \tilde{\theta}_{[0:N-1]}, \tilde{u}^H_{[0:N-1]})$, we refer to the corresponding state and belief state trajectory $(\tilde{x}_{[0:N]},\tilde{\bel}_{[0:N]})$ as a \emph{scenario}.
Note that by expanding \eqref{eq:sto_DP} using \eqref{eq:exp_expand}, the total number of scenarios is  $(|\tilde{\Xi}||\tilde{\ctrlset}^H|)^N$.
As a result, the optimization problem can quickly become intractable due to an exponentially growing number of decision variables.
Therefore, we use ST-SMPC to solve the problem over a subset of representative human uncertainty realizations.
\begin{remark}
\label{rmk:responsive_H}
Recall that in \eqref{eq:Boltzmann} we use a human's state-action value function $Q^H_{\theta_t} (x_t, u_t^H)$ that depends on the robot's state $x^R_t$.
This introduces coupling between uncertainties and decision variables in \eqref{eq:exp_expand}, which significantly increases the complexity of the optimization.
In order to plan in real time, we consider a class of human state-action value function parametrized as $Q^H_{\theta_t} (x^H_t, u_t^H)$, which (conservatively) assumes that the human does not react to the robot.
As a result, the human's action model \eqref{eq:Boltzmann} equals $P(u_t^H \mid x_t^H, \beta_t, \theta_t)$.
Nonetheless, we show in Section \ref{sec:experiment} that our method is still effective with a ``responsive'' human, whose unmodeled responses cause a reduction in the inferred inverse temperature $\beta_t$, similar to~\cite{fisac2018probabilistically}.
Our work may be extended to
explicitly account for human reactions leveraging recent advances in dual SMPC with state-dependent uncertainty~\cite{bonzanini2020safe}.
\end{remark}

\subsubsection{Constructing a sparse scenario tree}
Our proposed scenario tree construction procedure is summarized in Alg.~\ref{alg:sparse_tree} and depicted in Fig.~\ref{fig:sparse_tree}.
We start by introducing some useful definitions.
We denote a \emph{node} in the tree as $n$, whose state and belief state are denoted as $x_n$ and $b_n$.
The set of all nodes is defined as $\cN$.
We define the transition probability from a parent node $\pre(n)$ to its child node $n$ as $\bar{P}_{n} := \sum_{(\beta,\theta) \sim b_{n}} b_n(\beta, \theta) \cdot P(u^{H} \mid \pre(x^H_n), \beta, \theta).$
Subsequently, the \textit{path transition probability} of node $n$, i.e. the transition probability from the root node $n_0$ to node $n$ can be computed recursively as $P_{n} := \bar{P}_{n} \cdot \bar{P}_{\pre(n)} \cdots \bar{P}_{n_0}$.

\begin{algorithm}[!htbp]
	\caption{Constructing a sparse LQG scenario tree}
	\label{alg:sparse_tree}
	\begin{algorithmic}[1]
	\Require Current state $x_t \in \Omega$ and belief state $b_t$, maximum number of nodes $M>0$, truncated horizon $\bar{N} \leq N$, surrogate policy $\pi_\QMDP(x,b)$
	\Ensure A scenario tree defined by node sets $\cN_t, \cN^s_t$
	\LineComment{Initialization:}
	\State $x_{n_0} \gets x_t$, $b_{n_0} \gets b_t$, $t_{n_0} \gets 0$, $P_{n_0} \gets 1$
	\State $\cN_t \gets \{n_0\}$, $\cN^s_t \gets \emptyset$, $m \gets 1$, $n_{\br} \gets n_0$
	\While{$m \leq M$ }
	    \LineComment{Forward Simulation for One Scenario:} \label{alg:sparse_tree:fwd_sim_begin}
	    \State $\tn \gets n_\br$
	    \ForAll{$k \gets t_{n_\br}, t_{n_\br}+1, \ldots, \bar{N}-1$}
	        \LineComment{Robot Control:}
	        \State {$u^R_{\tn} \gets \pi_\QMDP(x_{\tn},b_{\tn})$}
	        \If{{$(x_{\tn},u^R_{\tn}) \in \shieldset^R$}} \Comment{Shielding required}
	            \State $u^R_{\tn} \gets \pi^s(x_{\tn})$
	            \State $\cN_t^s \gets \cN_t^s \cup \{\tn\}$
	        \EndIf
	        \LineComment{Human Control:}
	        \If{$k=t_{n_\br}$ \begin{bf}and\end{bf} $|\cN_t|>1$ } \Comment{Branching}
	            \State $u^H_{\tn} \gets u^H_\br$
	        \Else   \Comment{Non-branching}
	            \State $u^H_{\tn}  \gets \argmax \sum_{\beta, \theta} b_{\tn}(\beta, \theta) P(u^{H} | x^H_{\tn}, \beta, \theta)$
	        \EndIf
	        \State Compute path transition probability: $P_{n_m} \gets P_{\tn} \cdot \sum_{\beta, \theta} b_{\tn}(\beta, \theta) \cdot P(u^H_{\tn} \mid x^H_{\tn}, \beta, \theta)$
	        \State Update state: $x_{n_m} \gets {f}\left(x_{\tn}, u^R_{\tn}, u^H_{\tn} \right)$
	        \State Update belief state: $b_{n_m} \gets g(b_{\tn}, x_{\tn}, u^H_{\tn})$ \label{alg:sparse_tree:belief_update}
    	    \State $\cN_t \gets \cN_t \cup \{n_m\}$, $\tn \gets n_m$, $m \gets m + 1$
	    \EndFor \label{alg:sparse_tree:fwd_sim_end}
	    \State $(n_\br, u^H_\br) \gets \textsc{GetBranchNode}(\cN_t)$ \label{alg:sparse_tree:branch}
	\EndWhile
	\State $\cN_t \gets \textsc{NormalizePathTransProb}(\cN_t)$ \label{alg:sparse_tree:norm_path_trans_prob}
	\end{algorithmic}
\end{algorithm}

In order to efficiently leverage belief state propagation for predicting future shielding events, our scenario tree construction procedure differs from the conventional ones~\cite{bernardini2011stabilizing,arcari2020approximate, bonzanini2020safe} in three key aspects.
First, at scenario branching time ({Alg.~\ref{alg:sparse_tree},} Line~\ref{alg:sparse_tree:branch}), we only need to draw samples for human's actions $u^H$ but not for the parameters $(\beta,\theta)$.
Importantly, those $u^H$ samples are only used for updating the belief states ({Alg.~\ref{alg:sparse_tree},} Line~\ref{alg:sparse_tree:belief_update}).
In the next section, we show that by exploiting the problem structure, the robot's action obtained by solving the SMPC will adapt to the belief states instead of the samples.
Second, after each scenario branching, instead of propagating the (belief) states for only one time step, we perform a forward simulation up to a truncated horizon of $\bar{N} \leq N$ ({Alg.~\ref{alg:sparse_tree},} Line~\ref{alg:sparse_tree:fwd_sim_begin}-\ref{alg:sparse_tree:fwd_sim_end}).
This generally leads to a \emph{sparse} scenario tree with an increased depth, allowing us to capture more shielding events in the future.
Finally, when branching out new nodes ({Alg.~\ref{alg:sparse_tree},} Line~\ref{alg:sparse_tree:branch}), instead of selecting nodes with higher {realization} probabilities~\cite{bernardini2011stabilizing}, we are interested in those that lead to \emph{distinct trajectories}, which are essentially shaped by different shielding events.
Concretely, 
\new{when picking a new branch node $n_\br$, we prioritize one with a smaller time step $t_\br$,
which is likelier to result in a distinct trajectory from the existing ones in the tree.}
At node $n_\br$, we sample several human's action $\tilde{u}^H \in \ctrlset^H$, each of which produces a scenario $(\tilde{x}_{[0:\bar{N}]}, \tilde{b}_{[0:\bar{N}]})$ via forward simulation.
We then pick $u^H_\br = \tilde{u}^H$ that leads to the most different scenario from all existing ones in the tree.
The difference between two scenarios is measured in terms of the difference in the metric ${\xi}^\top H \xi$, where $\xi$ is a vector stacking all components of $\tilde{x}_{[0:\bar{N}]}, \tilde{b}_{[0:\bar{N}]}$ and $H$ is a positive semidefinite matrix.

\subsubsection{Optimizing over LQG scenarios}
In ST-SMPC, given a scenario tree, one shall optimize simultaneously for each scenario a robot's action sequence, which reacts to the human uncertainty in that scenario.
One key difference of our approach from ST-SMPC literature~\cite{arcari2020approximate,bernardini2011stabilizing,bonzanini2020safe} is that the optimized robot's action $u^R_\tn$ at node $\tn$ does not react to the \emph{samples}, i.e. the human's action $u^H_\tn$, but to the entire \emph{distributions} $P(u^H \mid x^H_\tn, \beta, \theta)$ and $b_\tn$.
Specifically, given a scenario $(x_{\tn_{[0:\bar{N}]}},b_{\tn_{[0:\bar{N}]}})$ associated with node sequence $\tn_{[0:\bar{N}]}$, the corresponding scenario optimization problem is,
\vspace{-1pt}
\begin{equation}
\label{eq:branch_opt}
\begin{aligned}
\min_{\bar{u}^{R}_{\tn_{[0:\bar{N}-1]}}} \ &\sum_{k=0}^{\bar{N}-1} \operatorname*{\mathbb{E}}\limits_{\substack{ (\beta, \theta) \sim b_{\tn_k}, \\
{u}_{\tn_k}^{H} \sim P(u^{H} \mid x^H_{\tn_k}, \beta, \theta )  } } \ell^R (\bar{x}_{\tn_k}, \bar{u}^R_{\tn_k})
\end{aligned}
\end{equation}
subject to constraints \eqref{eq:SHARP:sys_init}, \eqref{eq:SHARP:sys_dyn} and \eqref{eq:SHARP:control}, where we use $\bar{(\cdot)}$ to denote decision variables.
If this scenario shares nodes with other scenarios (e.g. node $n_0$ in Fig.~\ref{fig:sparse_tree}), then the robot's action at those shared nodes should be constrained to be the same, which enforces \emph{causality}~\cite{bernardini2011stabilizing}. 

One key observation of \eqref{eq:branch_opt} is that if $Q^H_{\theta}(x^H, u^H)$ is approximated as a quadratic function of $u^H$, then the human's action uncertainty $P ( u^{H} \mid x^H, \beta, \theta )$ becomes a Gaussian distribution with mean $\hat{u}^H(\beta,\theta) := \argmax P(u^{H} \mid x^H, \beta, \theta)$.
Furthermore, we linearize the joint dynamics \eqref{eq:joint_sys} around scenario trajectories $x_{\tn_{[0:\bar{N}]}}$, $u^R_{\tn_{[0:\bar{N}]}}$ and $u^H_{\tn_{[0:\bar{N}]}}$ to obtain a linear dynamical system,
\begin{equation}
\label{eq:linearized_sys}
\dx^+= A_{\tn_k} \dx + B^R_{\tn_k} \du^R + B^H_{\tn_k} \du^H_{\tn_k},
\end{equation}
where $\dx = x - x_{\tn_k}$, $\du^R = u^R - u^R_{\tn_k}$, $\du^H_{\tn_k} = \hat{u}^H_{\tn_k} - u^H_{\tn_k}$ and $A_{\tn_k}$ is the Jacobian $D_{x_{\tn_k}} f(\cdot)$, likewise for $B^R_{\tn_k}$ and $B^H_{\tn_k}$.
If we further drop the shielding constraint \eqref{eq:SHARP:control} for a moment (we will return to this in the next section) and consider a quadratic cost $\ell^R$, then \eqref{eq:branch_opt} becomes a Linear-Quadratic-Gaussian (LQG) problem, whose optimal solution is known to be certainty-equivalent~\cite{athans1971role}.
The resulting robot's control sequence $u^{R}_{\tn_{[0:\bar{N}-1]}}$ will be robust to distributions $P(u^H \mid x^H_{\tn_k}, \beta, \theta)$ and $b_{\tn_k}(\beta, \theta)$.

\subsubsection{Convexifying the shielding constraint}
\label{sec:SHARP:CBF}
\begin{figure}[!hbtp]
  \centering
  \includegraphics[width=0.45\columnwidth]{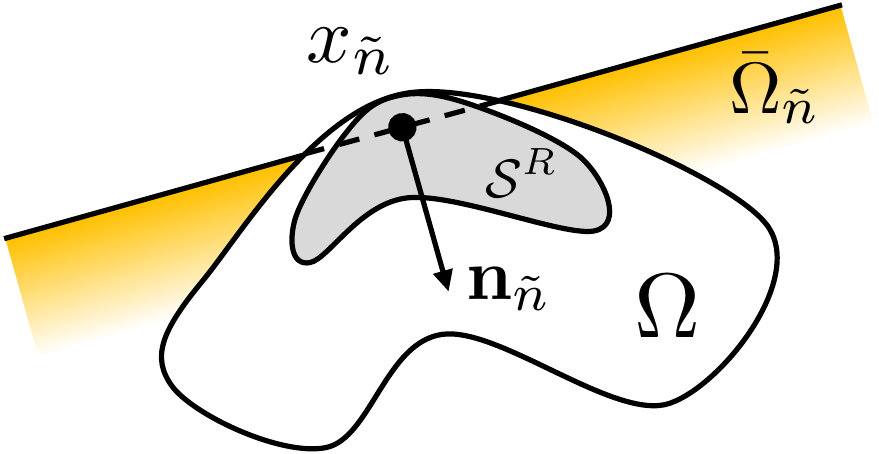}
  \caption{\label{fig:CBF} Illustration of a CBF-based convex shielding constraint.
  }
\end{figure}

The final piece we need to deal with is the shielding constraint \eqref{eq:SHARP:control}, which is in general non-convex.
In this paper, we propose to convexify it using the discrete-time exponential control barrier function (CBF) developed in \cite{agrawal2017discrete}.
The main idea is to linearize the system and approximate the safe set as a halfspace at any state $x \in \shieldset^R_{(\cdot)}$, in which case an affine CBF can be constructed analytically~\cite{agrawal2017discrete}.
Concretely, given a shielding node $\tn \in \cN^s$, we first obtain a linearized system at $(x_{\tn}, u_{\tn})$ according to \eqref{eq:linearized_sys}.
We then approximate the safe set $\Omega$ locally at $x_\tn$ as a halfspace defined by
\begin{equation}
\label{eq:safe_set_CBF}
\bar{\Omega}_\tn := \{ x \mid {\bn_\tn}^\top (x - x_\tn) \geq 0 \} = \{ \dx \mid {\bn_\tn}^\top \dx \geq 0 \},
\end{equation}
where $\bn_\tn := f(x_\tn, \pi^s(x_\tn), u^H_\tn) - x_\tn$ approximates the normal vector of the tangent space of $\shieldset^R_{u^R_{\tn}}$ at $x_\tn$, as illustrated in Fig.~\ref{fig:CBF}.

\begin{proposition}
\label{prop:CBF}
\textit{\cite[Prop. 4]{agrawal2017discrete}}
Given a safe set $\bar{\Omega}_\tn$ define by \eqref{eq:safe_set_CBF}, the affine function $h_\tn(\dx) := {\bn_\tn}^\top \dx$ is a discrete-time exponential CBF for system \eqref{eq:linearized_sys} linearized at $(x_{\tn}, u_{\tn})$ if there exists $\gamma \in (0,1]$ and $u^R \in \ctrlset^R$ such that $\forall \dx \in \bar{\Omega}_\tn$, $h_\tn\left( A_{\tn} \dx + B^R_{\tn} \du^R + B^H_{\tn} \du^H \right)+(\gamma-1)h_\tn (\dx) \geq 0$ holds.
\end{proposition}

Using the CBF defined in Proposition \ref{prop:CBF}, we can now approximate the shielding constraint \eqref{eq:SHARP:control} as,
\begin{equation}
\label{eq:convex_shielding_constraint}
   \bn_\tn^\top \left[ \left(A_\tn + (\gamma-1)I \right) \dx + B^R_\tn \du^R + B^H_\tn \du^H \right] \geq 0,
\end{equation}
which is linear (and hence convex) in $\dx$ and $\du^R$.

\begin{remark}
{As pointed out in~\cite{agrawal2017discrete}, constraint \eqref{eq:convex_shielding_constraint} is not necessarily feasible for bounded control input. Therefore, we incorporate it as a soft constraint in the ST-SMPC problem.}
\end{remark}


\subsubsection{Overall ST-SMPC Problem for SHARP}
\label{sec:SHARP:ST-SMPC}
Given a sparse LQG scenario tree defined by node sets $\cN_t$ and $\cN^s_t$, we can approximate \eqref{eq:sto_DP} as an ST-SMPC problem, 
\begin{equation}
\label{eq:ST-SMPC}
\begin{aligned}
\min_{\substack{\Pi_t}} \ \ 
&\sum_{\tn \in \cN_t \setminus \cL_t} \sum_{\beta, \theta} b_\tn(\beta, \theta) P_{\tn} \ell^R (\bar{x}^{\beta,\theta}_\tn, \pi_\tn) \\
&+ \sum_{\tn \in \cL_t} \sum_{\beta, \theta} b_\tn(\beta, \theta) P_{\tn} V_F (\bar{x}^{\beta,\theta}_\tn, b_\tn) \\
\text{s.t.} \ \ &\forall \tn \in \cN_t \setminus \cL_t: \pi_\tn \in \ctrlset^{R}, \\
&\forall \tn \in \cN_t \setminus \{n_0\}: \eqref{eq:linearized_sys}, \\
&\forall \tn \in \cN^s_t: \eqref{eq:convex_shielding_constraint},
\end{aligned}
\end{equation}
where $\cL_t$ is the set of all leaf nodes $\tn$ with $t_\tn = \bar{N}$, $\Pi_t := \{ \pi_\tn(\bar{x}^{\beta,\theta}_\tn,b_\tn): \tn \in \cN_t \setminus \cL_t \}$ is the collection of robot's control inputs associated with all non-leaf nodes, and $V_F(\cdot,\cdot)$ is the QMDP value function defined in \eqref{eq:QMDP_ter_cost}.
The path transition probabilities are normalized such that they sum up to $1$ at each time step (Alg.~\ref{alg:sparse_tree}, Line~\ref{alg:sparse_tree:norm_path_trans_prob}).
Problem \eqref{eq:ST-SMPC} is a quadratic program and thus can be solved efficiently.
The optimal solution $\Pi_t^*$ to \eqref{eq:ST-SMPC} is implemented in a receding horizon fashion, i.e. $\pi_{\text{SMPC}}(x_t, b_t) = \pi_{n_0}^{*}$.
We refer to this policy as SHARP-SMPC.

\section{Results}
\label{sec:experiment}
In this section, we evaluate SHARP on simulated driving scenarios, where we use the human driver's trajectories both from the Waymo Open Motion Dataset~\cite{sun2020scalability} and simulated using a car-following model in~\cite{krauss1998microscopic}.
{For simulation purposes}, vehicle dynamics are described by a {kinematic} bicycle model~\cite{fisac2019hierarchical} and discretized with a time step of $\Delta t = 0.2$ s;
{for planning, we use the linearized model from the Running Example}.
All simulations are performed using MATLAB and YALMIP~\cite{Lofberg2004yalmip} on a laptop with an Intel Core i7-7820HQ CPU.
The code and dataset are available at \url{https://github.com/SafeRoboticsLab/SHARP}

\textbf{Ablation.} We consider an ablation method that uses the state-of-the-art stochastic MPC scheme~\cite{arcari2020approximate}, which is based on the ST-SMPC technique originally developed in~\cite{bernardini2011stabilizing}, but additionally propagates the belief states that allows for human motion prediction via~\eqref{eq:Boltzmann} and \eqref{eq:belief_state_dyn}.
The MPC only has control constraints.
Therefore, the scenario information is only used by the objective function and the resulting policy is \emph{safety-unaware} {(though nonetheless \emph{safe} thanks to shielding)}.

\textbf{Baseline.} Our baseline method adds to the ablation soft collision-avoidance constraints of $x_\tn \notin \F, \ \forall \tn \in \cN$.
We used {a simple} grid search to determine {approximately} optimal weights for the soft constraints.
Note that the baseline policy is \emph{safety-aware} but \emph{shielding-agnostic}.

\begin{figure}[!hbtp]
  \centering
  \includegraphics[width=1.0\columnwidth]{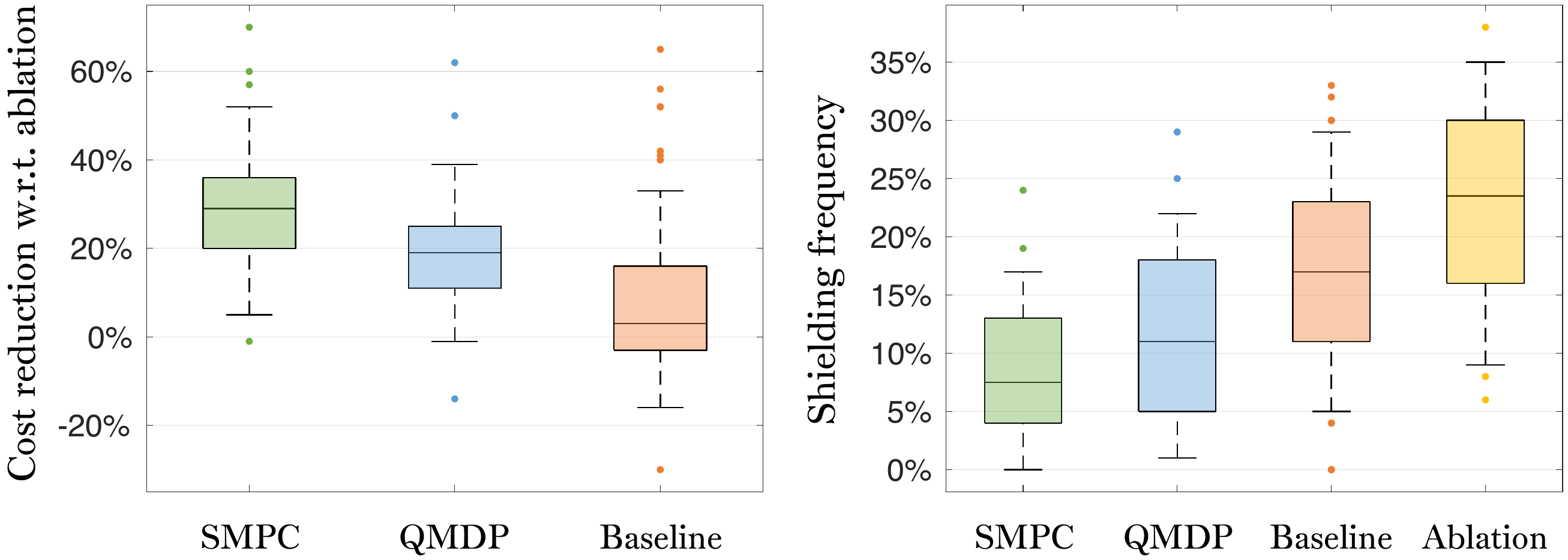}
  \caption{\label{fig:Sc1_Stats} Cost reduction and shielding frequency of Scenario 1 with 50 human trajectories from the Waymo Open Motion Dataset~\cite{sun2020scalability}. The central mark, bottom and top edges of the box indicate the median, 25th and 75th percentiles, respectively. The max whisker length is the interquartile range. Outliers are shown as points.}
\end{figure}

\textbf{Simulation Setup.} The ablation, baseline and SHARP planners are equipped with the same HJ-reachability-based shielding policy~\cite{bansal2017hamilton}.
They also use the same human intent inference scheme~\eqref{eq:Boltzmann} and \eqref{eq:belief_state_dyn} to obtain a prediction of human's future trajectories.
All ST-SMPC problems use a bound $M=70$ on the number of nodes in the tree, and are solved with MOSEK~\cite{aps2019mosek} (average solving time 60 ms).

\textbf{Metrics.}
 We first define the closed-loop cost as $J^R_\cl := \sum_{t=0}^{T_\Sim} \ell^R(x_t, u_t^R)$, where $T_\Sim$ is the simulation horizon, and $x_{[0:T_\Sim]}, u_{[0:T_\Sim]}$ are the \emph{executed} state and input trajectories (with replanning).
To measure the performance of the planners, we consider the following two metrics:
\begin{itemize}
    \item Cost reduction rate: Defined as the percentage reduction of the closed-loop cost achieved by a certain planner with respect to the one achieved by the ablation.
    \item Shielding frequency: A number defined as $T_\shield / T_\Sim \times 100\%$, where $T_\shield$ is the number of time steps when shielding is used.
\end{itemize}

\subsection{Scenario 1: Highway Overtaking}
We first show simulation results for Scenario 1, which is the running example.
We simulate the scenario for 50 times, each with a different human's trajectory taken from the Waymo Open Motion Dataset~\cite{sun2020scalability}.
The performance metrics are presented in Fig.~\ref{fig:Sc1_Stats}.
We observe that SHARP planners outperform the baseline in both metrics, due to their ability to take advantage of human inference to predict the costly shielding events.
On the other hand, even though the baseline also leverages human inference for collision avoidance, the heuristic proximity penalty can negatively interfere with the robot's actual performance criterion, and is ultimately less effective at preventing unnecessary shielding events.

\begin{figure}[!hbtp]
  \centering
  \includegraphics[width=1.0\columnwidth]{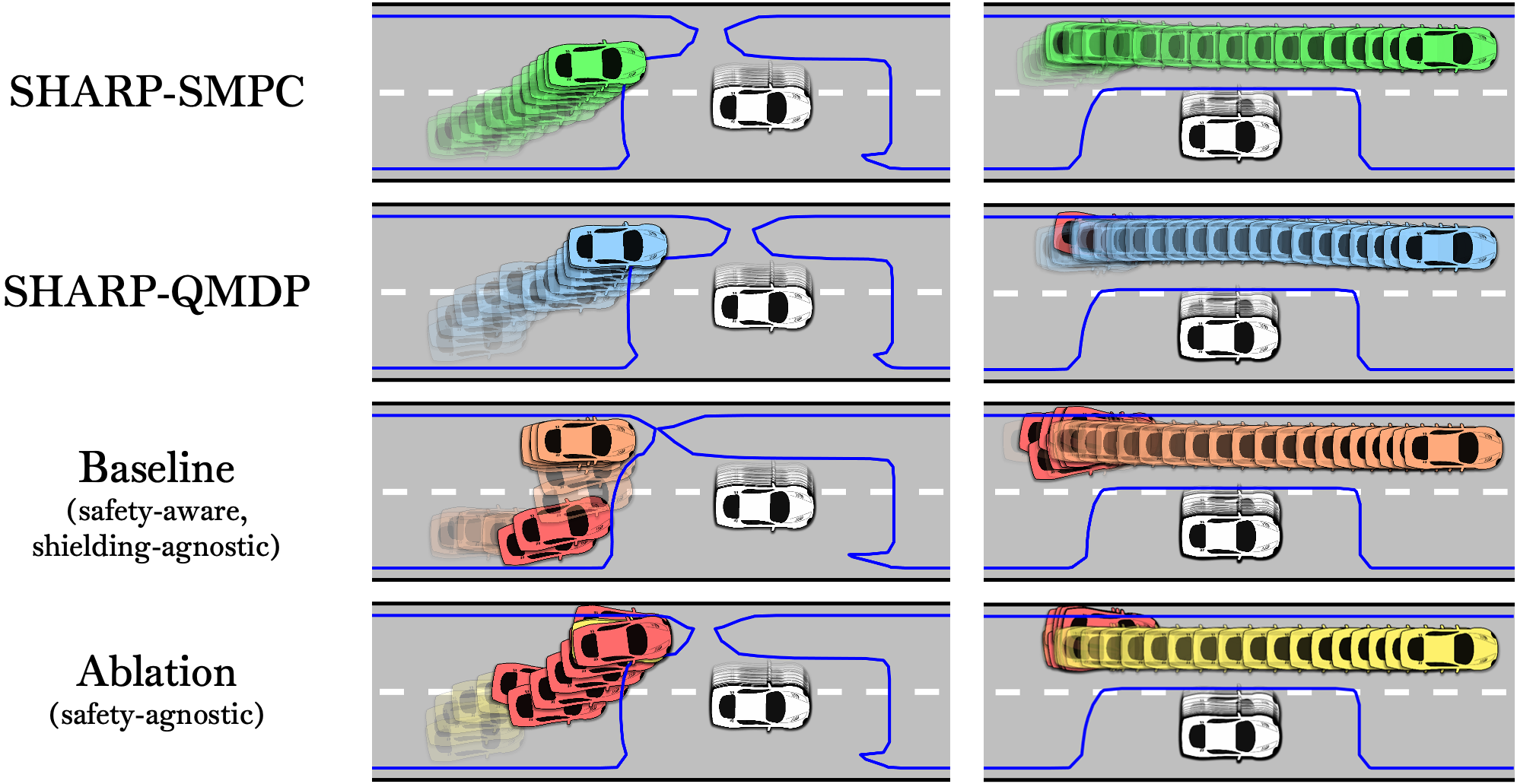}
  \caption{\label{fig:Sc1_Traj} Simulation snapshots of Scenario 1. Longitudinal positions are shown in relative coordinates with $p_x^H = 0$. The left column displays trajectories for $t=[0,4.8]$ s and the right one displays the remainder of the trajectories. $(p_x^R,p_y^R)$-slices of the safe set $\Omega$, taken at the terminal state in each trajectory, are indicated in blue. A red vehicle snapshot indicates a shielding override.}
\end{figure}

Snapshots of one simulation trial are shown in Fig.~\ref{fig:Sc1_Traj}.
We observe that SHARP-SMPC accurately predicts the human's future movement to the right lane, controls the robot to stay in the left lane, following the human without incurring shielding (top left), and safely overtakes the human when a window of opportunity opens (top right).
SHARP-QMDP, despite rendering a low shielding frequency as well thanks to the shielding-awareness, cannot as  effectively reason about and react to the human's uncertain trajectory due to the overly optimistic QMDP assumption, resulting in a more conservative trajectory.
The baseline triggers more shielding events and produces a less efficient trajectory than the SHARP planners due to lack of shielding-awareness.

\subsection{Scenario 2: Traffic Intersection}
\begin{figure*}[!hbtp]
  \centering
  \includegraphics[width=1.35\columnwidth]{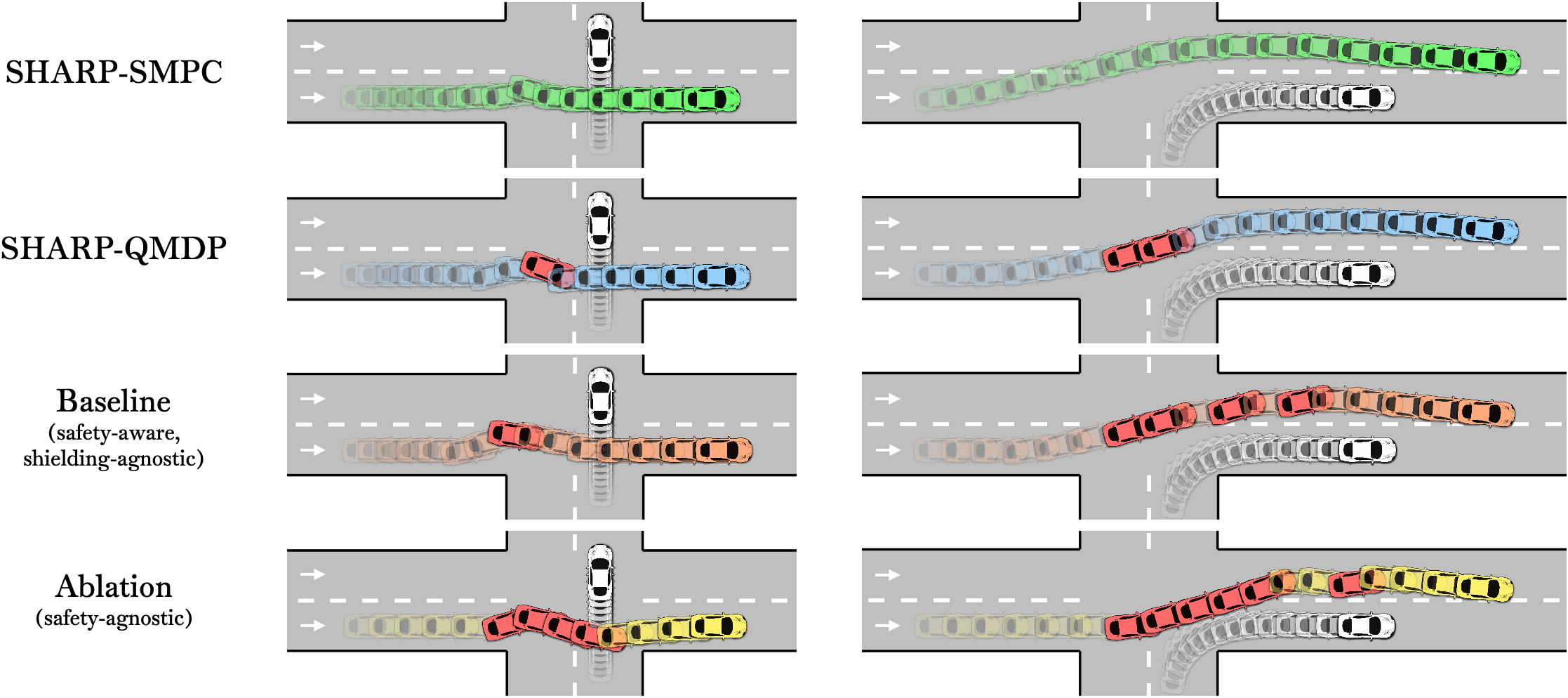}
  \caption{\label{fig:Sc2_Traj} Simulation snapshots of Scenario 2. A red vehicle snapshot indicates a shielding override.}
\end{figure*}

\begin{figure}[!hbtp]
  \centering
  \includegraphics[width=0.96\columnwidth]{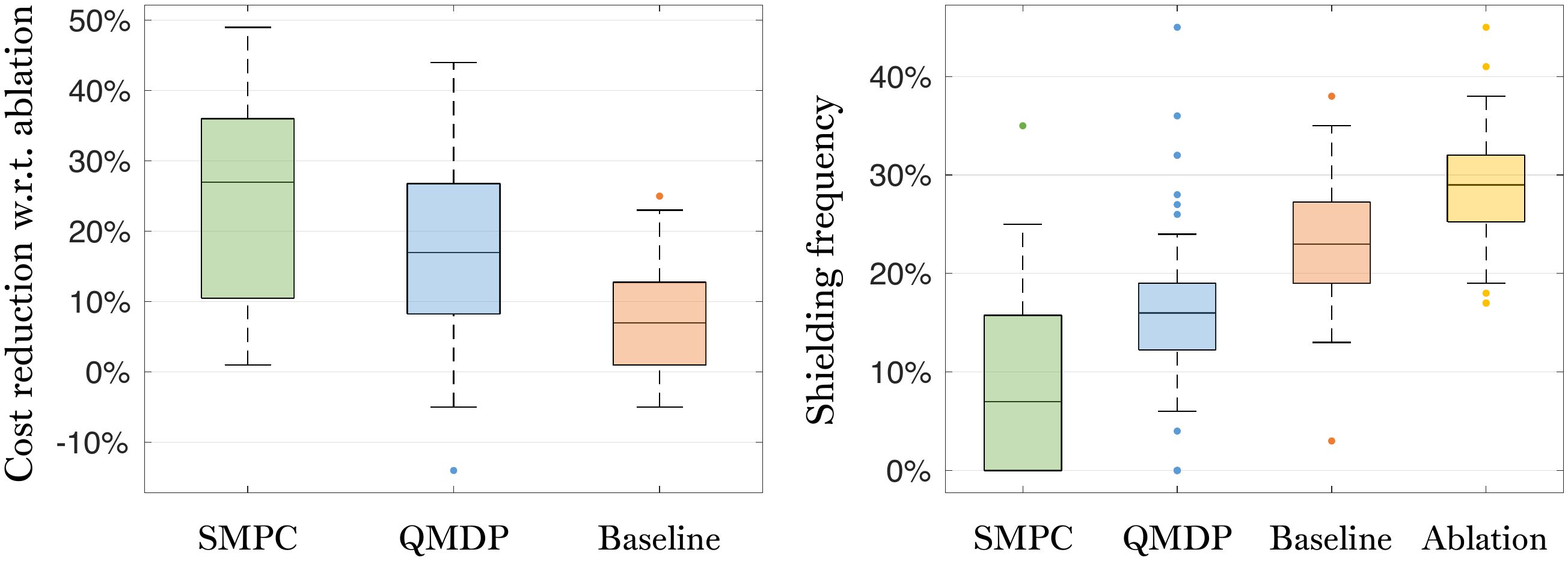}
  \caption{\label{fig:Sc2_Stats} Cost reduction and shielding frequency of Scenario 2 with 50 human trajectories from the Waymo Open Motion Dataset~\cite{sun2020scalability}.}
\end{figure}

Next, we consider a traffic intersection scenario where the human may choose to stop, go straight or make a right turn.
The performance metrics obtained from 50 simulation trials with human's trajectories taken from the Waymo Open Motion Dataset are shown in Fig.~\ref{fig:Sc2_Stats}.
Snapshots of two simulation trials with the human going straight and turning right are depicted in Fig.~\ref{fig:Sc2_Traj}.

\subsection{Responsive Human}
\begin{figure}[!hbtp]
  \centering
  \includegraphics[width=0.96\columnwidth]{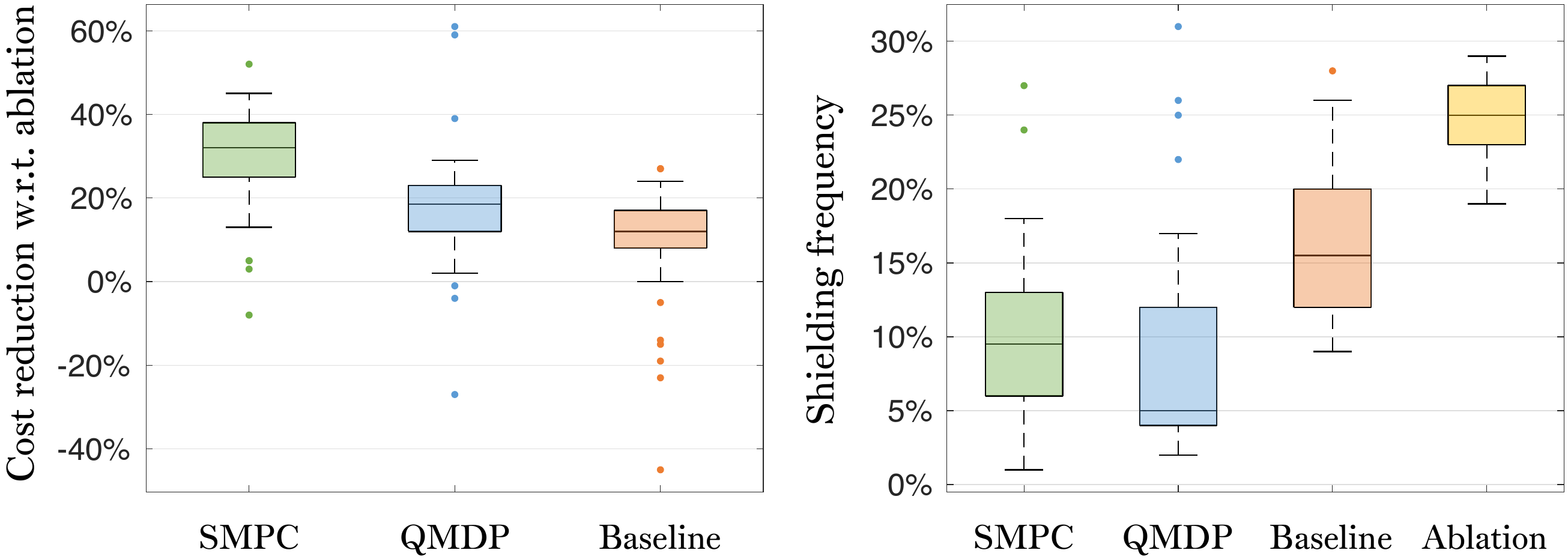}
  \caption{\label{fig:RH_Stats} Cost reduction and shielding frequency of Scenario 1 obtained from 50 trials, where the human is simulated using~\cite{krauss1998microscopic}.}
\end{figure}

Finally, we revisit Scenario 1 with a responsive human (see Remark \ref{rmk:responsive_H}).
We simulate the behaviour of the human with the car following model from~\cite[Chapter~4]{krauss1998microscopic}, which is also used in microscopic traffic simulators such as SUMO~\cite{SUMO2018}.
The parameter values we used are human's preferred acceleration $a = 3$ m/s$^2$, reaction time $\tau = 1$ s, and random velocity perturbation $\eta = 0.1$ m/s.
The human also performs random lane changing maneuvers.
The performance metrics obtained from 50 simulation trials are shown in Fig.~\ref{fig:RH_Stats}.
We observe that even {in the face of} unmodeled human {behavior}, SHARP planners still outperform the baseline.

\section{Discussion} 
\label{sec:conclusion}
\looseness=-1
\noindent \textbf{Summary.} We have introduced Shielding-Aware Robust Planning (SHARP), a decision-making framework for safe and efficient interaction.
The SHARP policy improves {robustness} by accounting for possible future shielding events,
{proactively balancing nominal performance with costly emergency maneuvers triggered by unlikely human behaviors.}

\looseness=-1
\noindent {\textbf{Limitations and future work.} Performance of SHARP policies can be limited by neglecting human reactions to the robot's future decisions.
The scenario tree approach provides a promising avenue for extended formulations that tractably account for human responsiveness.
Similarly, scalability improvements are needed in order to compute real-time SHARP policies for multi-human multi-robot interaction.
Finally, the current framework assumes that the robot can accurately observe the state and past human actions, which is often unrealistic.
Combining the efficient risk mitigation of SHARP with
recent advances in safe perception-aware planning~\cite{zhang2021safe} is likely to yield more general and powerful frameworks.}


\balance
\printbibliography
\end{document}

\end{document}